# Synthetic Fluency: Hallucinations, Confabulations, and the Creation of Irish Words in LLM-Generated Translations

**Sheila Castilho**
SALIS / ADAPT Centre
Dublin City University
sheila.castilho@dcu.ie

***Zoe Fitzsimmons, Claire Holton, Aoife Mc Donagh**†
SALIS, Dublin City University
first.second2@mail.dcu.ie
†first.second75@mail.dcu.ie

## Abstract

This study examines hallucinations in Large Language Model (LLM) translations into Irish, specifically focusing on instances where the models generate novel, non-existent words. We classify these hallucinations within verb and noun categories, identifying six distinct patterns among the latter. Additionally, we analyse whether these hallucinations adhere to Irish morphological rules and what linguistic tendencies they exhibit. Our findings show that while both GPT-4.o and GPT-4.o Mini produce similar types of hallucinations, the Mini model generates them at a significantly higher frequency. Beyond classification, the discussion raises speculative questions about the implications of these hallucinations for the Irish language. Rather than seeking definitive answers, we offer food for thought regarding the increasing use of LLMs and their potential role in shaping Irish vocabulary and linguistic evolution. We aim to prompt discussion on how such technologies might influence language over time, particularly in the context of low-resource, morphologically rich languages.

## 1 Introduction

Since the emergence of neural machine translation (MT), hallucinations have been recognised as a significant challenge in the field (Koehn and Knowles, 2017). Large Language Models (LLM), which have shown great capabilities for translation, also produce hallucinations, and, despite efforts by both the MT and natural language processing communities to identify and address these issues, they remain prevalent, particularly in low-resource scenarios (Sennrich et al., 2024). Hallucinations produced by LLMs are claimed to be "qualitatively different from those of conventional translation models" which include "off target translations, overgeneration, and even failed attempts to translate" (Guerreiro et al., 2023, p. 1501).

Building on these challenges, this study focuses on specific types of hallucinations, namely instances where the system *invents* new words during translation. Our goals are to identify which word classes are affected when open LLMs generate hallucinations in a low-resource language like Irish (Gaeilge) and to assess whether these hallucinated words follow Irish linguistic rules or diverge from established conventions. To the best of our knowledge, this is the first study to examine the morphology of hallucinations generated when translating into Irish.

## 2 Background

### 2.1 Hallucinations

Recent advancements in artificial intelligence, particularly the rise of decoder-only LLMs like GPT-3.5 and GPT-4, have ushered in a new era for MT (Brown et al., 2020). These models are renowned for their exceptional performance in understanding and generating human language due to their capacity to learn from massive datasets and produce high-quality translations (Hendy et al., 2023; Moslem et al., 2023).

Despite their impressive capabilities, generative LLMs continue to face significant difficulties when translating low-resource languages (Castilho et al., 2023; Robinson et al., 2023). These challenges arise from the severe under-representation of low-resource languages in available training data. As a result, even when various prompts are used, the translations produced often reflect "poor generalization" and may be "inaccurate or nonsensical" due to the models' "limited exposure to the linguistic nuances" of these languages (Shu et al., 2024).

One of the issues LLMs face is that of halluci-nations (Bang et al., 2023). Several works have recorded the types of hallucination that LLMs produce in different NLP tasks (Ji et al., 2023; Huang et al., 2024). In the context of narrative and dialogue generation, Sui et al. (2024) suggest that hallucination are not necessarily "inherently harmful" and may offer potential benefits, referring to them as "confabulations."

⁰* These authors contributed equally to this paper and are named alphabetically.

Few studies have addressed hallucination in LLM-based MT (Guerreiro et al., 2023; Sennrich et al., 2024; Briakou et al., 2024), with most focusing on the detection and mitigation. Our interest lies in examining the morphology of specific types of hallucinations and confabulations, particularly those involving the creation of entirely new words by the systems.

### 2.2 Irish Morphology

Examining the morphology of invented hallucinations is especially relevant for the Irish language, as its morphological structure relies heavily on the use of suffixes, infixes and prefixes to form plurals for nouns, the genitive case for nouns and to conjugate verbs and for the formation of new words (Cassidy, 2024). Like other morphologically rich languages, Irish exhibits a high degree of inflection and relatively free word order, which poses specific challenges for MT from English (Cotterell et al., 2018). These structural differences often lead to errors in translation output, particularly when models trained predominantly on English struggle to accurately generate complex morphological forms or correctly interpret flexible syntactic structures (Lankford et al., 2021). Cotterell et al. (2018) suggest that morphological typology may explain some of the variability in model performance across languages, noting that LLMs tend to perform worse on highly inflected languages. Similarly, Arnett and Bergen (2025) provide evidence of a performance gap between languages of different morphological types, raising broader concerns about unintended disparities in how languages are treated in NLP systems. Given these findings, investigating hallucinations in Irish—a morphologically rich but low-resource language—may provide insights into broader trends affecting similar languages.

For instance, in Irish, **nouns** are categorised into five declensions (see Appendix A), where "the defining criterion for each admission to each category is the form of the genitive singular ending" (Ball and Muller, 2010, p.177). The construction of the plural form in Irish consist of two categories, '*lagiolraí*' (weak plurals) and the '*treaniolraí*' (strong plurals). Weak plurals are mainly found in the first and second declensions. In the first declension, plural formation typically involves palatalisation of the final consonant, whereas in the second declension, many nouns form their plurals by adding -a to the singular form. Strong plurals encompass all other plural formations, as nouns in the 3rd, 4th and 5th declensions take strong plural endings. Examples of such endings include *-(e)acha, -(e)annna, -(a)í, -t(h)a and -t(h)e*.

Similarly, **verbs** have the addition of initial mutations, such as lenitions and eclipses. Each tense and mood has its own unique set of suffixes for the conjugation of verbs (see Appendix B). Irish verbs are formed by classification into two conjugations: the first and second conjugations (*an chéad réimniú agus an dara réimniú*). The first conjugation consists of "all one-syllable verbs, two-syllable verbs ending in *-(e)áil* and a small number of two-syllable verbs, which are not syncopated (lose their second syllable) when a third or fourth syllable is added" (Ball and Muller, 2010, p.189). The second conjugation is comprised of all other two-syllable verbs. Within the first and second conjugations, there are two possible suffixes depending on the type of vowels in the roots. **Broad vowels** (*leathan*) *-a, -o, -u* must be followed by the suffix beginning with a broad vowel; and **Slender vowels** (*caol*) *-i, -e* must be followed by the suffix beginning with a slender vowel. A *lenition* is used to mark the past and imperfect tenses, the conditional mood and also follows the negative particles, the conjunction *má* and the interrogative particle *ar*. A lenition is also used following the direct relative clause particle *-a*.[1]

Additionally, since the seventeenth century, Irish has been influenced almost entirely by the English language with "the most dramatic changes have occurred in the last 100 years, in the period when the monolingual Irish speaker became a rarity" (Hickey, 2009, p. 671). As such, there is a tendency to borrow lexicon from English, and adapt these borrowings to align with its grammatical and morphological rules, known as lexical borrowing with adaptation (Mulhall, 2018). Similarly, "new loans replacing existing Irish words", which has been referred to as 'detrimental change' has been noted in recent years (Hickey, 2009, 671). Once such example is the case of the word 'zoo', which appears in de Bhaldraithe's 1959 English-Irish Dictionary as *gairdín ainmhithe*.[2] In Ó Dónaill's 1977 *Foclóir Gaeilge-Béarla* (Irish-English Dictionary)[3] the word 'zoo' appears as *zú*, while the previous translation *gairdín ainmhithe* (garden of animals) is no longer listed. In parallel, code-switching—defined as "instances of the linguistic phenomenon that results in mixed-language text" (Lynn and Scannell, 2019, p.33) — has also become increasingly common in contemporary Irish usage (Cassidy, 2024).

---
[1] Table 16 in the Appendix, shows an example of how there are four possible categories for suffixes when conjugating verbs in most Irish tenses and moods.

### 2.3 Automatic Translation of Irish

Due to its intricate morphology described above, not to mention the flexible word order, and rich inflectional system, the Irish language poses significant challenges when translating from English. These challenges are even more pronounced for automatic systems, where maintaining grammatical accuracy in features such as noun gender and case inflections proves particularly difficult (Lankford et al., 2023). Nonetheless, the challenge of MT for Irish has been documented in several research works (Dowling et al., 2018, 2020; Lankford et al., 2021).

Regarding the development of LLMs for Irish, we highlight the work of Lankford et al. (2023), who fine-tuned multilingual models for translating low-resource languages, including Irish. Additionally, Tran et al. (2024b) report on a pioneering effort to develop an open-source Irish-based LLM, proposing a framework to adapt an English-centric model into a bilingual system. Their results demonstrate strong performance in both understanding and generating Irish text; however, challenges remain, particularly issues such as the forgetting of English as "a consequence of continued pre-training on Irish data" (Tran et al., 2024a, p.194).

Given the above, a better understanding of the models' ability to handle the complexities of low-resource languages like Irish is necessary. To achieve that, we aim to analyse whether hallucinated words generated by LLMs conform to Irish morphological rules or diverge entirely. To frame our analysis, we draw on the definitions of hallucinations proposed by Huang et al. (2024), who classify them into *factuality hallucination* and *faithfulness hallucination*, where the former is a discrepancy on verifiable real-world fact, and the latter "captures the divergence of generated content from user input or the lack of self-consistency within the generated content" (ibid, p.42:2). Moreover, *faithfulness hallucination* is subdivided into *context inconsistencies*, which arise when generated content misaligns with the provided context.[4] Under this definition, the phenomenon of the model inventing new words falls within the category of faithfulness hallucinations, specifically as context inconsistencies.

The term "*confabulation*" has been proposed as a more accurate alternative to hallucination. Sui et al. (2024) argue that "LLM confabulations mirror a human propensity to utilize increased narrativity as a cognitive resource for sense-making and communication" (p.14274). They define confabulation as a narrative-driven tendency to organise available information into coherent stories, even when key details are missing—leading to the generation of plausible yet fictional content. From this viewpoint, the model's invention of new words that resemble legitimate Irish morphology can be framed as confabulations. Therefore, in this paper, we use *hallucination* as a general term to refer to all outputs that diverge from the source content or expected translation, while we reserve the term *confabulation* for hallucinated outputs that invent new words which appear internally coherent and plausible according to Irish morphological rules.

These definitions provide a foundation for analysing the morphological patterns observed in LLM-generated outputs when translating from English into Irish, allowing us to distinguish between different types of invented words and their potential implications.

## 3 Experimental Setup

### 3.1 Test Set

In order to evaluate these types of hallucinations LLMs generate when translating into Irish, we conducted a preliminary pilot test. We identified that general texts (such as general news from the WMT corpora) did not produce any of these hallucinations. However, domain-specific texts, particularly those in scientific and medical fields containing a higher frequency of unfamiliar terms, showed noticeable examples of these hallucinations. Based on these observations, we selected three texts from the DELA corpus (Castilho et al., 2021)[5] for this experiment: two scientific news texts and one technical TED Talk, as shown in table 1.

---

[2] https://www.teanglann.ie/ga/eid/zoo
[3] https://www.teanglann.ie/ga/fgb/zoo

[4] The other two categories, "instruction inconsistency" (where content deviates from the original instruction) and "logical inconsistency" (where internal contradictions occur), are not relevant to the type of hallucination studied here.

| Document Title | Domain | # Sentences | # Tokens |
|---|---|---|---|
| Giant fans of wind energy | News | 55 | 4898 |
| Arm processors: Everything you need to know now | News | 40 | 5691 |
| On the verge of creating synthetic life | TED Talk | 130 | 13605 |

Table 1: Test Set Statistics.

We note that our test sets were published before 2022, making it likely that the models have encountered them during training. However, this does not pose an issue for our research since the terminology used in the test set is intentionally chosen, as it is more likely to trigger hallucinations, as observed in our pilot study. This aligns with our objective of testing the models' performance in handling challenging, domain-specific content in Irish.

### 3.2 LLMs

The pilot phase involved testing three open LLMs: ChatGPT 4.0,[6] Co-Pilot, and Gemini.[7] However, both Co-Pilot and Gemini presented significant challenges, as their outputs were notably verbose (Briakou et al., 2024), even after multiple attempts to refine the translation process, with many refusals to translate.

Due to these limitations, we decided to focus on two versions of ChatGPT: 4.o (henceforth, GPT4) and 4.o Mini (henceforth, Mini). It should be noted that users accessing ChatGPT 4.o are switched to the Mini version after exceeding the limit of 50 messages within a 3-hour period.[8]

#### 3.2.1 Prompts

Mizrahi et al. (2024, p.935) warn against the limitations of single-instruction evaluation of LLMs, claiming that "a simple rephrasing of the instruction, template can lead to drastic changes in both absolute and relative model performance". We note however that, since our goal is for the LLMs to produce hallucinations in order to analyse the construction of those, we opted for a simple prompt to translate the source and not to give any comments on the output (Sennrich et al., 2024).

**Prompt:** *Translate this text from English into Irish. Translate all words except named entities, and just respond with the translation, without any additional comments:* [full source text]

If the output contained untranslated words, we followed up with a secondary prompt to address the issue:

**Follow-up Prompt:** *The word(s) [untranslated word(s)] was/were not translated. Retranslate the full text making sure to translate these words. Just respond with the full text translation without any additional comments.*

Full texts were given so the model could make use of the whole context.[9]

## 4 Detecting and Analysing Hallucinations in Irish

As noted previously, some characteristics of the Irish language, such as the heavy reliance on the use of suffixes and prefixes, and the great number of compound words, pose a great challenge for automatic translation. After the translation of the test sets, we observed a significant number of hallucinations related to verbs and nouns, adverbs and with fewer involving adjectives. Due to space and time constraints, we focus on hallucinations related to verbs and nouns. Table 2 presents the frequency of hallucinations across all test sets for both GPT4 and Mini.

We note that the number of invented hallucinated words is greater for *nouns*, with only a few instances for *verbs*. Moreover, the Mini model shows a greater number of invented hallucinated words in comparison with GPT4, showing a rate of 2.14 hallucinations of this type, against 0.86 hallucinations for the latter. This is an expected result regarding the model's performance, since Mini is a smaller and less robust version of GPT4. Smaller models generally have fewer parameters, which can impact their ability to accurately handle complex linguistic phenomena, such as Irish morphology and inflection. Nonetheless, since our objective is not to compare the models' outputs but rather to analyse the patterns in which these hallucinations are generated, differences in the number of hallucinations, as well as variations in model architecture and size, do not impact the validity of our study.

---

[5] https://github.com/SheilaCastilho/DELA-Project
[6] https://chatgpt.com
[7] see copilot.microsoft.com and gemini.google.com
[8] https://help.openai.com/en/articles/9275245-using-chatgpt-s-free-tier-faq?utm_source=chatgpt.com

[9] We note that due to the Mini model's tendency to stop translating when presented with longer texts, we divided the input of test set 3 into three segments. Each segment consistently included the beginning of the text, ensuring that key details such as the talk's title, the speaker's name, and relevant keywords were preserved.

| Model | Verb | Noun | Total | Rate |
|---|---|---|---|---|
| GPT4 | 06 | 15 | 21 | 0.86 |
| Mini | 04 | 48 | 52 | 2.14 |
| Total | 10 | 63 | 73 | - |

Table 2: Frequency of hallucinations involving invented words across all test sets for each model. A normalised hallucination rate is expressed as the number of hallucinations per 1,000 tokens.

| Model | Rules | No Rules | Total | % Rules |
|---|---|---|---|---|
| GPT4 | 04 | 02 | 06 | 67 |
| Mini | 02 | 02 | 04 | 50 |

Table 3: Frequency of hallucinations related to *VERBS* across all test sets. "Rules" indicates hallucinations that followed Irish grammatical and morphological rules (confabulations), while "No Rules" refers to those that did not adhere to these rules.

### 4.1 Hallucinating Irish Verbs

Table 3 presents the total number of hallucinated verbs and indicates whether they adhere to Irish grammatical and morphological rules. From the six invented hallucinated verbs by GPT4, four of them follow the Irish rules for grammar and morphology, and are classified as confabulations. Their application in context are shown in table 4.

We observe that when GPT4 confabulates verbs, its most common strategy is to reinterpret the source verb (e.g., 'sequenced', 'code', 'sequence') as a noun and then generate a corresponding Irish word. This results in the invention of forms such as *shraitheamar*, *códálann*, and *shraitheadh*, which, if they were actual Irish verbs, would be morphologically well-formed.

For example, in Example 1 in table 4, GPT4 has taken the noun *sraith*, meaning 'sequence' or 'series', and has correctly added the first person plural slender conjugation in the past tense, and lenitised the verb correctly, as is required in the past tense. In Example 2, GPT4 has taken the noun *cód* (which means 'code') and conjugated it using the correct broad present tense ending. However, it has added an additional syllable *ál*, which seems to align with the convention of verbs such as *tástáil* ('to test') which is conjugated as *tastálann* in the present tense. In Example 3, GPT4 has again taken the noun *sraith*, as in Example 1, and conjugated it into the past tense autonomous verb, the *briathar saor*. It has correctly lenitised the verb, as the direct relative clause particle 'a' proceeds it.

Example 4 shows another common type of confabulated verb. In this case, GPT4 adopts a well-documented feature of the Irish language — borrowing (Mulhall, 2018) words from English — while retaining the original English spelling and attempting to 'conjugate' them according to Irish grammatical patterns. *Tendeann* results from GPT4 taking the English verb 'tend' and correctly conjugates it into the first conjugation ending for slender vowels. There is no singular equivalent in Irish to the English verb 'tends to'.

While we decided that the listed examples are technically morphological, GPT4 also generated hallucinations that were not morphologically sound. For example, the verbal noun *athsraitheadh*. Here, the prefix *ath* (similar to 're-' in English) was applied to express the repetition of an action. However, while a lenition should typically follow a prefix in the stem of the verb, GPT4 omitted this. Another example of unnecessary omissions included the hallucination, *chog*. While seemingly attempting to translate the verb 'to chew', GPT4 omitted the latter half of '*chogain*' from its infinitive form and conjugated it into the first conjugation.

Regarding invented hallucinated verbs from the Mini model, from the four reported in table 2, two of them follow the Irish rules for grammar and morphology and are shown in table 5.

Similar to GPT4, the Mini model also generates confabulated verbs that follow two main patterns: transforming a source-language verb into a target-language noun, which then conjugated as if it were a verb, or retaining an English word while conjugating it according to Irish morphological rules. Example 1 in Table 5 illustrates the conjugation of the English verb 'simulate', by removing the third syllable and adding the correct present tense suffix *-aíonn*. Example 2 shows the conjugations of an Irish noun *cód* ('code') which has been used as the root of the verb and had a correct present tense ending of the first conjugation for broad vowels applied.

| Verbs | source | output | type |
|---|---|---|---|
| 1 | When we first sequenced this genome | Nuair a **shraitheamar** an géanóm seo ar dtús | conjugation of a noun |
| 2 | Triplets of those letters code for roughly 20 | **Códálann** tripléid de na litreacha sin do thart ar 20 | conjugation of a noun |
| 3 | so we could sequence them .. | go bhféadfaimis iad a **shraitheadh** | conjugation of a noun |
| 4 | Each device incorporating an Arm processor tends to be | **Tendeann** gach gléas a chuimsíonn próiseálaí Arm a bheith | English word conjugated |

Table 4: Confabulated verbs that followed the Irish morphology rules by GPT4.o.

| Verbs | source | output | type |
|---|---|---|---|
| 1 | it doesn't simulate the execution of code | nach **simulaíonn** sé comhoibriú cód | English word conjugated |
| 2 | Triplets of those letters code for roughly 20 | **Códann** trípéirí de na litreacha sin thart ar 20 | Conjugation of a noun |

Table 5: Confabulated verbs that followed the Irish morphology rules by GPT4.o. Mini

| Model | Rules | No Rules | Total | % |
|---|---|---|---|---|
| GPT-4.0 | 11 | 04 | 15 | 73 |
| Mini | 19 | 29 | 48 | 39 |

Table 6: Frequency of hallucinations related to *NOUNS* across all test sets. "Rules" indicates hallucinations that followed Irish grammatical and morphological rules (confabulations), "No Rules" refers to those that did not adhere to these rules, and "Total" represents the overall number of hallucinations for each model.

Invented hallucinated verbs which did not follow the rules were : *dearthach* which was the translation given for 'designing'. It appears that the model mistook 'designing' for an adjective and tried to translate it as that. The root *dear*, 'design' is correct, but in the second syllable it seems the model has combined the verbal adjective *deartha* and the suffix *-ach*, which commonly features in Irish adjectives.

*Aknowimid* was the translation given for 'you know' (human translation: *tá a fhios agat/agaibh*) in the source text. *Aknowimid* uses the incorrect root, given that the Irish alphabet does not feature the letter 'k', and has been conjugated incorrectly using the slender first conjugation rather than the broad second conjugation. It seems that the model has attempted to say 'we acknowledge' even though it deviates slightly from the source to avoid a phrase that it was unfamiliar with.

## 4.2 Hallucinating Irish Nouns

As previously shown, the majority of hallucinated words in Irish were nouns. This is unsurprising given the intricacies of the five declensions of Irish nouns (see Appendix B). Table 6 presents the total number of hallucinated nouns generated by both models and indicates whether they adhere to Irish grammatical and morphological rules. To better structure the analysis of these hallucinated nouns, this section is divided into the following types: Compounds (section 4.2.1), Lazy Gaelicisation (section 4.2.2), Good Hallucination (section 4.2.3), Code-switching (section 4.2.4), Prefix (section 4.2.5), and Suffix(section 4.2.6).

### 4.2.1 Compounds

We note that both models have used a compounding of nouns to create invented hallucinated words. Table 7 illustrates the one instance of compounding of two nouns in GPT4 *bhinncheisteanna* which compounds the noun *binn* ('peak', 'cliff' or 'edge') and *ceisteanna* ('questions').

In Irish, compounding often involves initial consonant mutations in the second or subsequent parts of the compound (Ball and Muller, 2010, 176). Therefore, the hallucinated word *bhinncheisteanna* follows this pattern correctly, applying lenition to the second component, *ch*eisteanna. No other compound nouns, either morphologically correct or incorrect, was invented by this model.

Regarding invented compounds by the Mini model, table 8 illustrates the 5 confabulated examples that could be classified as morphologically correct, although they carry little meaning.

Example 1 *gaothmhoill* (attempted translation of 'windmill') is a compounding of the word *gaoth* ('wind') and *moill* ('delay'). The morphological rule of initial consonant mutations (lenitions) is followed. A pattern emerged in the hallucinations created for this category in Mini, whereby, the first noun in the compound is correct or relates to the source text, but is followed by an incorrectly translated noun. The second noun *moill* is nonsensical in this context, however, it does resemble the English noun 'mill'.

Example 2 *gaothchumhachta* (attempted translation of 'turbine and wind turbine') compounds *gaoth* ('wind') and *cumhacht* ('power'). This translation differs greatly from the human translation *tuirbín* and *tuirbín gaoithe*. A lenition is applied correctly to the second noun and it is correctly in the genitive singular in all 9 cases, as is required.

|   | source | GPT4.o |
|---|---|---|
| 1 | ...results of independent performance benchmarks... | ...torthaí de **bhinncheisteanna** feidhmíochta neamhspleácha... |

Table 7: Confabulated **Compound Nouns** that followed the Irish morphology rules by GPT4.o

|   | source | Mini |
|---|---|---|
| 1 | Or, in this case, windmill. | Nó, sa chás seo, **gaothmhoill**. |
| 2 | Evolution of the turbine | Evoláid na **gaothchumhachta** |
| 3 | ...modern wind turbines are huge... | ...tá **gaothmhoillí** nua-aimseartha ollmhóra... |
| 4 | Wind turbines are reaching ever higher. | Tá **gaothchumhachtaí** ag dul níos airde agus níos airde. |
| 5 | results of independent performance benchmarks | torthaí na **gcomhairlíon** próiseálaí neamhspleácha |

Table 8: Confabulated **Compound Nouns** that followed the Irish morphology rules by GPT4.o Mini.

|   | source | GPT4.o |
|---|---|---|
| 1 | ...on all of the elements in the nacelle. | ...ceann de na heiliminití sa **nascáil**. |
| 2 | Triplets of those letters code for roughly | ódálann **tripléid** de na litreacha sin do thart ar |
| 3 | ...what we're calling combinatorial genomics | ...atá á ghlaoch againn **géanómóireacht** chomhcheangailteach |

Table 9: Confabulated words that followed the Irish morphology rules by the GPT4.o classified as '**Lazy Gaelicisation**".

|   | source | Mini |
|---|---|---|
| 1 | It will handle turbine blades... | Rachaidh sé i ngleic le **blaide** gaothchumhachta... |
| 2 | so we thought we'd build them in cassettes... | mar sin shocraíomar iad a thógáil i **gcásáidí**... |
| 3 | ...this may sound like genomic alchemy... | ...b'fhéidir go mbeidh sé seo cosúil le **alcaimíocht** ghéineamach... |
| 4 | Now I've argued, this is not genesis; | Anois, rinne mé argóint, ní **ghinéise** atá anseo; |

Table 10: Confabulated words that followed the Irish morphology rules by the GPT4.o Mini classified as '**Lazy Gaelicisation**".

Example 3 *gaothmhoillí* is similar to example 1, but the second noun is in the plural. However, the word to be translated in Example 1 is 'windmill', in contrast to 'wind turbines' in Example 3.

Example 4 is similar to Example 2, as it also compounds *gaoth* and *cumhacht*, however, the second noun is in the nominative and genitive plural, which is correct in all 6 cases.

Example 5 shows *comhairlíon* as a translation for 'benchmarks'. It compounds *comhair* ('combined work', 'co-operation', 'partnership'), with *líon* (a full number, complement). It is morphologically correct, as it follows orthographic rules (broad vowels followed by broad vowels, slender vowels followed by slender). A lenition is not added to the second word, as a lenition cannot be added to an 'l'. Example 5 shows less logic than the other pattern and seems to compound two random nouns to create an invented hallucination.

The only invented hallucination for compounded nouns by the Mini model that did not follow morphological rules was *gaoithchumachta*, which while similar Example 2 in Table 8 contains and 'i' in *gaoth*, meaning it does not follow orthographic rules.

### 4.2.2 Lazy Gaelicisation

We refer to instances where translations appear to have been generated based on the phonetics of the English word, often modifying the spelling to conform to Irish orthographic rules even though a corresponding word exists in Irish as *Lazy Gaelicisation*. Both engines (GPT4 in table 9, and the Mini model in table 10) produced confabulations of this variety. Many of these confabulated words could plausibly be mistaken for legitimate Irish terms, particularly in casual reading. At the very least, the reader would recognise their connection to the English source and infer the intended meaning with relative ease. These phonetic adaptations have been found among Irish speakers, particularly in informal or spontaneous speech, and sometimes in writing (Darcy, 2014). The GPT4 model presented a few of those cases. We note that these examples represent a clear alignment with our definition of confabulation: invented words that are not simply erroneous, but which exhibit internal coherence and plausibility according to Irish phonological and morphological norms.

|   | source | GPT4 |
|---|--------|------|
| 1 | ...heart of a device controller, a microcontroller (MCU) | ...chroílár rialtóra gléas, **micririaltóir** (MCU) |
| 2 | this is just a regular photomicrograph. | níl anseo ach **fótamhicreagraf** rialta. |
| 3 | ...with synthetic bacteria, Archaea... | ...le baictéir shintéiseacha, **Seanríochtaí**... |

Table 11: 'Good' Confabulated words that followed the Irish morphology rules by GPT4.o

|   | source | Mini |
|---|--------|------|
| 1 | Giant fans of wind energy | **Fanaithe** ollmhóra de fuinneamh gaoth |
| 2 | ...in what sources outside of Apple call an "emulator" | ...ar a dtugtar "**simulachtóir**"... |
| 3 | ...invention, science, technology | ...**inventiú**, eolaíocht, teicneolaíocht |

Table 12: Confabulated words with **code-switching**, that is, English nouns that followed the Irish morphology rules for by GPT4.o Mini.

In Example 1 in table 9, the model translated the word 'nacelle' as *nascáil* ('linkage') instead of *naoisil*, which is the correct translation. We note that both terms ('nacelle' and *nascáil*) are extremely phonetically similar which could explain this hallucination. Moreover, 'nacelle' is highly specialised language relating to aeronautical engineering, and therefore, it is entirely possible that the Irish term is newly coined, after the model was last updated as the national terminology database for Irish is updated constantly.[10]

In Example 2 in table 9, GPT4 translated 'triplets' as *tripléid* (correct translation is *tríríní*), in which the second syllable *-pléid* demonstrates a correct pluralisation. 'Triplet' and *tripléid*, which we assume the model believes is the singular, are phonetically similar justifying the model's reasoning. In example 3, GPT4 translated 'genomics' as *géanómóireacht* (correct translation is *géanómaíocht*). The model has added an unnecessary syllable, however the reasoning is unclear.

The Mini model also presented a few examples of *Lazy Gaelicisation* as shown in table 10. The model translated 'blades' as *blaide* (correct translation is *lanna*), following the orthographical rules by matching the slender vowels. In example 2, the model translated 'cassettes' as *cásáidí* (correct translation is *caiséid*). Irish nouns ending in *-áid* are usually feminine, belonging to the 2nd declension. Therefore, they are pluralised using the suffix *-í*, as the model has done. The Irish noun *caiséad* is masculine in the 1st declension, meaning the last consonant must be slenderised to produce the plural (both nominative and genitive case).

Example 3 shows the translation of the word 'alchemy' as *alcaimíocht* (correct translation is *ailceimice*). The Irish suffix *-(a)íocht* is commonly used to express the English suffix '-ation'. For example, *reachtaíocht*, 'legislation', *eagraíocht*, 'organisation', *radaíocht*, 'radiation', *cúrsaíocht*, 'circulation'(of money), *cáilíocht*, 'qualification'. It is possible that the model took the context of the test set into consideration and was influenced by domain-specific nouns that it was familiar with. In example 4, the model translated the term 'genesis' as *ginéise*. The intended meaning in the source text refers to the beginning of something, therefore *bunús* is used to express this in Irish.

Examples in this category from the Mini model that do not follow morphological rules include: *protáitíopaíocht* (human translation *fréamhshamhaltú*) and *protáitíopaí* (human translation *fréamhshamhlacha*), used to translate the terms 'prototyping' and 'prototypes' respectively, used the incorrect prefix for 'proto-'; *evólúisian* as the translation for 'evolution', disregards the convention of the Irish alphabet which does not include the letter 'v'; *autagrafaí*, used to translate 'autograph', uses the incorrect prefix for 'auto-' which is usually *uath-*. In this case, the correct translation is *átagraf*.

### 4.2.3 Good Confabulations

We classify good confabulations as invented outputs that seem to follow all morphological rules for words which had no official translation available, but a good attempt has been made to create a word. These cases demonstrate creative yet coherent language generation in the absence of concrete lexical data. There were 3 such cases produced by GPT4 (see table 11), and no cases by the Mini model.

---

[10] https://www.tearma.ie/

|   | source | Mini |
|---|--------|------|
| 1 | mainly in the area of composites. | go príomha i réimse na **gcomhshamhlacha**. |
| 2 | Triplets of those letters code for roughly 20 amino acids, | Códann **trípéirí** de na litreacha sin thart ar 20 aigéad aimín, |
| 3 | we think that biology can have a major impact | gur féidir leis an **bithleacht** níos mó tionchar a imirt |

Table 13: Confabulated words that followed the Irish morphology rules for **prefix** by GPT4.o Mini.

|   | source | GPT4 |
|---|--------|------|
| 1 | ...there's a problem when it comes to simulating wind turbines. | ...tá fadhb ann maidir le **turasáin** gaoithe a insamhladh. |
| 2 | ...forces and moments on the shaft in three directions. | ...fórsaí agus **cuimhneachtaí** ar an seafta i dtrí threoir. |
| 3 | that can take three million rads of radiation. | is féidir a ghlacadh trí mhilliún **radaim** radaíochta. |
| 4 | Archaea and, eventually, eukaryotes. | Seanríochtaí agus, faoi dheireadh, **eocaróitigh**. |

Table 14: Confabulated words that followed the Irish morphology rules for **suffix** by GPT4.o.

|   | source | Mini |
|---|--------|------|
| 1 | starting with the digital information of the genome of phi X174. | ag tosú leis an eolas digiteach de **ghéineomaí** phi X174. |
| 2 | that can take three millions rads of radiation. | atá in ann trí mhilliún **radán** de radaíocht a ghlacadh. |
| 3 | we can select for viability... | is féidir linn roghnú le haghaidh **feidhmeannaíochta**... |

Table 15: Confabulated words that followed the Irish morphology rules for **suffix** by GPT4.o Mini.

Example 1 shows *micririaltóir* as a translation of 'microcontroller' and correctly compounds the prefix *micri-* with the noun *rialtóir* (person). While this is a good attempt, the correct 'controller' in this context would be *rialtán* ('switch', 'button', 'dial'). There is no lenition added following the prefix, as lenitions cannot be added to the letter 'r'. Example 2, a translation of 'photomicrograph' shows a similar pattern to Example 1. The prefix *fóta-* is correctly added to the noun *micreagraf*, and a lenition is correctly added following the prefix. Example 3, *Seanríochtaí* as a translation for Archea is interesting, as it compounds the adjective *sean* ('old') with the noun *ríochtaí* ('kingdoms'). This is of interest as it appears to use an understanding of *Archea* as the adjective 'archaic' and translates it as such to *sean*.

**Deceiving 'good hallucinations'-** are invented hallucinations which, similar to *Lazy Gaelicisation*, seem and sound like correct Irish words, but upon further inspection, carry no meaning. This is the case of the word *laigeas* (produced by the Mini model), in an attempt to translate 'bending moments' from the source text while *Laigeas* appears to be a morphologically correct word, it contains no real units of meaning.[11]

### 4.2.4 Code-switching

We look into examples where the models have taken an English noun and added an Irish suffix in an attempt to create an Irish word. This phenomena has been reported in the use of Irish in tweets and been classified as code-switching word-level alternation (Lynn and Scannell, 2019). These examples differ from Lazy Gaelicisation in that they appear to be a compounding between the source language and the target language, disregarding the orthographical conventions of the Irish language. They also illustrate another facet of confabulation, where the system fills lexical gaps by improvising plausible word forms, albeit in ways that stretch or break conventional language norms. There were no occurrences of English nouns with Irish suffixes in GPT4.

In table 12 Example 1 *fanaithe*, the Mini model has taken the English noun 'fan' and added the Irish suffix *-aithe* which is commonly used to pluralise broad weak plural Irish nouns. In Example 2, *simulachtóir*, the model took the first two syllables of the noun 'simulator' and added the Irish suffix used to express '-ator', *-achtóir*. Regardless this is a mistranslation as the source calls for 'emulator'.

In Example 3, an attempt to translate the word 'invention' into *inventiú* was made by taking the first two syllables of the noun and adding the Irish suffix *-iú*.

The following examples within the code-switching category do not follow Irish morphological rules: *Simuláid*, which was an attempted translation of 'simulation'. In this case, the root 'simul' is not morphologically acceptable, the suffix *-áid* is seen across Irish in other nouns such as *cumarsáid* ('conversa**tion**') and *oráid* ('ora**tion**').

---
[11] source: 'able to withstand bending moments up to 100.000 kNm", output: '...atá in ann **laigeas** a fhulaingt suas le 100.000 kNm'.

Similarly, another example of this that was produced, is *evoláid*, the root of which 'evol' disregards the Irish alphabet which does not feature the letter 'v', it also suffixes *-áid*.

### 4.2.5 Prefix

Table 13 shows examples of hallucinations in which the Mini model created nouns using the correct prefixes long established within Irish morphology. In Examples 2 and 3, the model appeared to recognise the prefixes in their source form and translated them to Irish without correctly translating the latter parts of the nouns. Example 1 shows an attempt to have a similar function of the meaning of the source noun. We note that GPT4 model did not hallucinate any words with a 'correct' prefix.

Both GPT4 and the Mini model confabulated nouns with prefixes that were phonetically similar but incorrectly spelt. For example, *micoplásma* instead of *míceaplasma*, *cilivata* instead of *cileavata*. In other instances, both systems created hallucinations by keeping the prefix in its source form and translating the rest of the noun. For example, *megavata* and *sub-aonadanna*.

### 4.2.6 Suffix

The following hallucinations were identified and characterised by their use of real Irish nouns and the addition of an infix or suffix for a certain purpose.

In table 14, all listed hallucinations generated by GPT4 are concerned with pluralised nouns. Examples 1 and 3 show the inclusion of an infix in order to pluralise nouns, while the hallucinations that occurred in Examples 2 and 4 applied a suffix. These confabulations are deemed morphologically correct as they are typical of Irish spelling and also respect the conversions set out in the declensions.

The Mini model (table 15) generated confabulations that show phonological similarities to their correct translation, however the addition of suffixes could only be deemed unnecessary. In Example 1, the model produced the suffix *-aí* in *ghéineomaí*, which is commonly used to indicate a particular person or job in Irish. A possible explanation for this is that the Mini model may have misunderstood the source 'genome' to be an agent, rather than an object. Examples 2 and 3 show hallucinations where the first parts of the noun are correct, however noun endings that are common within Irish morphology were added. Interestingly, despite their incorrect endings, these nouns still respect the grammatical rules that are involved when counting items and turning a noun into its genitive case form.

Both systems also generated hallucinations where an apparent disconnect occurred between their spelling and patterns of mutations. For example, while attempting to translate 'voltage dips', GPT4 generated *dippaí*. This was deemed morphologically incorrect as it took the source noun, which is an existing loan word in Irish, that does not differ in the singular for the English 'dip', however, the correct plural is *dipeanna*. In this case GPT4 added double consonants (pp) and a strong plural ending. Similarly, the Mini model created the hallucination *titimeanna* for the same source phrase. The Mini model created incoherent hallucinations such as *dhearadhóir* in an attempt to translate 'designer'. The model took the Irish *dearadh*, meaning 'design' and attached a suffix that offers the same function as '-er'(*-óir*) in English to suggest an agent. This hallucination, however, was not deemed morphologically correct as it does not align with spelling conventions.

## 5 Discussion and Conclusions

This study examined the types of hallucinations involving the creation of new words in LLM-generated translations into Irish and evaluated whether these hallucinations adhered to Irish linguistic rules, and therefore classified as confabulations. Our findings indicate that both GPT4 and Mini exhibit similar patterns of word invention, though the latter produces hallucinated words at a significantly higher frequency. Specifically, the Mini model generated 52 hallucinated words compared to 21 from GPT4. While both models demonstrate a tendency to confabulate, that is, apply Irish morphological rules to these hallucinated words, GPT4 adheres to these rules more consistently (71%) than the Mini model (40%) (Tables 3 and 6). This difference likely reflects the Mini model's smaller size and reduced robustness. Nonetheless, both models produce *plausible* but *non-existent* lexical items that raise intriguing questions about their potential influence of confabulations on the Irish language.

One striking observation is that many of the confabulations resemble patterns made by learners of Irish, such as code switching and borrowings. This suggests that the models might not be generating entirely arbitrary forms but are instead applying Irish word formation rules in a way that mirrors natural language learning processes. These confab-ulation patterns are particularly relevant in the context of what Fhlannchadha and Hickey (2018, p.21) describe as a 'post-traditional variety of Irish'—a variety adopted by non-native speakers who do not align with any particular dialect of Irish. The authors note that established ideologies rooted in native and traditional models of Irish are being disrupted by new speakers, creating a notable tension between linguistic groups in the era of language revitalisation. They caution that the expansion of post-traditional Irish could lead to the erosion of crucial aspects of the language, particularly in lexicon and grammar. Similar concerns arise in other morphologically rich, low-resource languages, such as Scottish Gaelic, and Welsh, where language change and revitalisation efforts interact with evolving speaker communities. In this light, LLM-generated confabulations raise further questions about the role of AI in reinforcing or reshaping these dynamics across such languages.

But what does it mean when an AI model exhibits patterns akin to human learners? Could these errors, if encountered frequently in machine-generated content, influence the way Irish is written or even spoken over time? Two particularly noteworthy categories of confabulations observed in this study, which we term 'Lazy Gaelicisation' and 'Good Confabulation', involve the adaptation of English words into Irish-like phonetics, often by modifying their spelling to align with Irish orthographic rules. This phenomenon is not exclusive to LLMs; similar strategies have been observed among Irish speakers themselves. The phonetic adaptation of English words into Irish structures has long been a feature of the language, seen both in historical borrowings and in contemporary informal speech. Does this suggest that such hallucinations are merely an extension of a natural linguistic process? Or should they be viewed as problematic, reinforcing patterns of language shift rather than supporting authentic Irish usage?

We showed that the models invented words that follow all morphological rules when no official translation is available ('good confabulations'). The introduction of novel, non-standard words could be seen as either a sign of language erosion or a potential source of linguistic innovation. While some have found the replacement of existing Irish words with English-derived forms as a form of 'detrimental change' (Hickey, 2009, 671), others see partial or non-standard Irish as a step toward broader engagement. As one Irish-language commentator puts it, "broken Irish is better than smart English [...] broken Irish is a step towards fluency, not the end of the line."[12] If LLM-generated forms gain traction, could they help fill lexical gaps in technical domains where Irish terminology is scarce? Or would they risk further undermining existing Irish vocabulary? These are not straightforward questions, and rather than offering definitive answers, they highlight the need for continued observation and discussion.

Finally, it is important to highlight the specific context in which these hallucinations and confabulations occur. In our study, most invented hallucinated words appeared in technical and specialised domains, where even fluent speakers may struggle with terminology. In more general texts, where Irish has a more established lexicon, the models produced fewer invented words, although overall grammatical accuracy and fluency remained an issue. This suggests that while hallucinations in LLM-generated translations may be concerning in certain contexts, their broader impact on Irish will likely depend on how these models are used and integrated into real-world workflows.

Future research should explore these issues further with larger datasets and more extensive replication resources, particularly regarding the long-term implications of LLM-assisted translation for minority languages. Moreover, future work should look into how speakers perceive and react to these hallucinations and confabulations.

Nonetheless, this work serves as a foundation for further investigations into the implications of LLM errors in morphologically rich, low-resource languages. Our goal is to encourage discussion on the long-term impact of these technologies on language, especially in the case of low-resource, morphologically rich languages.

## Acknowledgments

We would like to thank Prof. Ciarán Mac Murchaidh for the invaluable help and discussions. The first author benefits from being member of the ADAPT SFI Research Centre at Dublin City University, funded by the Science Foundation Ireland under Grant Agreement No. 13/RC/2106_P2.

---

[12] https://tinyurl.com/bdemuvv4

# References

Catherine Arnett and Benjamin Bergen. 2025. Why do language models perform worse for morphologically complex languages? In *Proceedings of the 31st International Conference on Computational Linguistics*, pages 6607–6623, Abu Dhabi, UAE. Association for Computational Linguistics.

Martin Ball and Nicole Muller. 2010. *The Celtic Languages*, 2nd edition. Taylor Francis, Hoboken.

Yejin Bang, Samuel Cahyawijaya, Nayeon Lee, Wenliang Dai, Dan Su, Bryan Wilie, Holy Lovenia, Ziwei Ji, Tiezheng Yu, Willy Chung, Quyet V. Do, Yan Xu, and Pascale Fung. 2023. A multitask, multilingual, multimodal evaluation of ChatGPT on reasoning, hallucination, and interactivity. In *Proceedings of the 13th International Joint Conference on Natural Language Processing and the 3rd Conference of the Asia-Pacific Chapter of the Association for Computational Linguistics (Volume 1: Long Papers)*, pages 675–718, Nusa Dua, Bali. Association for Computational Linguistics.

Eleftheria Briakou, Zhongtao Liu, Colin Cherry, and Markus Freitag. 2024. On the implications of verbose llm outputs: A case study in translation evaluation. *Preprint*, arXiv:2410.00863.

Tom B. Brown, Benjamin Mann, Nick Ryder, Melanie Subbiah, Jared Kaplan, Prafulla Dhariwal, Arvind Neelakantan, Pranav Shyam, Girish Sastry, Amanda Askell, Sandhini Agarwal, Ariel Herbert-Voss, Gretchen Krueger, Tom Henighan, Rewon Child, Aditya Ramesh, Daniel M. Ziegler, Jeffrey Wu, Clemens Winter, Christopher Hesse, Mark Chen, Eric Sigler, Mateusz Litwin, Scott Gray, Benjamin Chess, Jack Clark, Christopher Berner, Sam McCandlish, Alec Radford, Ilya Sutskever, and Dario Amodei. 2020. Language Models are Few-Shot Learners. *arXiv preprint*. Issue: arXiv:2005.14165 arXiv:2005.14165 [cs].

Lauren Cassidy. 2024. *Linguistic analysis and automatic dependency parsing of Tweets in modern Irish*. Phd thesis, Dublin City University.

Sheila Castilho, João Lucas Cavalheiro Camargo, Miguel Menezes, and Andy Way. 2021. DELA Corpus - A Document-Level Corpus Annotated with Context-Related Issues. In *Proceedings of the Sixth Conference on Machine Translation*, pages 571–582. Association for Computational Linguistics (ACL).

Sheila Castilho, Clodagh Quinn Mallon, Rahel Meister, and Shengya Yue. 2023. Do online machine translation systems care for context? what about a GPT model? In *Proceedings of the 24th Annual Conference of the European Association for Machine Translation*, pages 393–417, Tampere, Finland. European Association for Machine Translation.

Ryan Cotterell, Sabrina J. Mielke, Jason Eisner, and Brian Roark. 2018. Are all languages equally hard to language-model? In *Proceedings of the 2018 Conference of the North American Chapter of the Association for Computational Linguistics: Human Language Technologies, Volume 2 (Short Papers)*, pages 536–541, New Orleans, Louisiana. Association for Computational Linguistics.

Guinevere Darcy. 2014. *Code-mixing and context: A Corca Dhuibhne case study*. Unpublished phd dissertation, University of Limerick.

Meghan Dowling, Teresa Lynn, Alberto Poncelas, and Andy Way. 2018. Smt versus nmt: Preliminary comparisons for irish. In *Proceedings of the AMTA 2018 Workshop on Technologies for MT of Low Resource Languages (LoResMT 2018)*, pages 12–20.

Meghan Dowling, Joss Moorkens, Andy Way, Sheila Castilho, and Teresa Lynn. 2020. A human evaluation of english-irish statistical and neural machine translation. In *22nd Annual Conference of the European Association for Machine Translation*, page 431.

Siobhán Nic Fhlannchadha and Tina M. Hickey. 2018. Minority language ownership and authority: perspectives of native speakers and new speakers. *International Journal of Bilingual Education and Bilingualism*, 21(1):38–53.

Nuno M. Guerreiro, Duarte M. Alves, Jonas Waldendorf, Barry Haddow, Alexandra Birch, Pierre Colombo, and André F. T. Martins. 2023. Hallucinations in large multilingual translation models. *Transactions of the Association for Computational Linguistics*, 11:1500–1517.

Amr Hendy, Mohamed Abdelrehim, Amr Sharaf, Vikas Raunak, Mohamed Gabr, Hitokazu Matsushita, Young Jin Kim, Mohamed Afify, and Hany Hassan Awadalla. 2023. How good are gpt models at machine translation? a comprehensive evaluation. *arXiv preprint*.

Tina Hickey. 2009. Code-switching and borrowing in irish. *Journal of Sociolinguistics*, 13(5):670–688.

Lei Huang, Weijiang Yu, Weitao Ma, Weihong Zhong, Zhangyin Feng, Haotian Wang, Qianglong Chen, Weihua Peng, Xiaocheng Feng, Bing Qin, and Ting Liu. 2024. A survey on hallucination in large language models: Principles, taxonomy, challenges, and open questions. *ACM Trans. Inf. Syst.* Just Accepted.

Ziwei Ji, Nayeon Lee, Rita Frieske, Tiezheng Yu, Dan Su, Yan Xu, Etsuko Ishii, Ye Jin Bang, Andrea Madotto, and Pascale Fung. 2023. Survey of hallucination in natural language generation. *ACM Comput. Surv.*, 55(12).

Philipp Koehn and Rebecca Knowles. 2017. Six challenges for neural machine translation. In *Proceedings of the First Workshop on Neural Machine Translation*, pages 28–39, Vancouver. Association for Computational Linguistics.

Séamus Lankford, Haithem Afli, and Andy Way. 2023. adaptmllm: Fine-tuning multilingual language models on low-resource languages with integrated llm playgrounds. *Information*, 14(12).

Séamus Lankford, Haithem Alfi, and Andy Way. 2021. Transformers for low-resource languages: Is féidir linn! In *Proceedings of Machine Translation Summit XVIII: Research Track*, pages 48–60, Virtual. Association for Machine Translation in the Americas.

Teresa Lynn and Kevin Scannell. 2019. Code-switching in Irish tweets: A preliminary analysis. In *Proceedings of the Celtic Language Technology Workshop*, pages 32–40, Dublin, Ireland. European Association for Machine Translation.

Moran Mizrahi, Guy Kaplan, Dan Malkin, Rotem Dror, Dafna Shahaf, and Gabriel Stanovsky. 2024. State of what art? a call for multi-prompt LLM evaluation. *Transactions of the Association for Computational Linguistics*, 12:933–949.

Yasmin Moslem, Rejwanul Haque, John D. Kelleher, and Andy Way. 2023. Adaptive machine translation with large language models. In *Proceedings of the 24th Annual Conference of the European Association for Machine Translation*, pages 227–237, Tampere, Finland. European Association for Machine Translation.

Chris Mulhall. 2018. Irish lexicography in borrowed time: The recording of anglo-irish borrowings in early twentieth-century irish dictionaries (1904-1927). *International Journal of Lexicography*, 31(2):214–228.

Nathaniel Robinson, Perez Ogayo, David R. Mortensen, and Graham Neubig. 2023. ChatGPT MT: Competitive for High- (but Not Low-) Resource Languages. In *Proceedings of the Eighth Conference on Machine Translation*, pages 392–418, Singapore. Association for Computational Linguistics.

Rico Sennrich, Jannis Vamvas, and Alireza Mohammadshahi. 2024. Mitigating hallucinations and off-target machine translation with source-contrastive and language-contrastive decoding. In *Proceedings of the 18th Conference of the European Chapter of the Association for Computational Linguistics (Volume 2: Short Papers)*, pages 21–33, St. Julian's, Malta. Association for Computational Linguistics.

Peng Shu, Junhao Chen, Zhengliang Liu, Hui Wang, Zihao Wu, Tianyang Zhong, Yiwei Li, Huaqin Zhao, Hanqi Jiang, Yi Pan, Yifan Zhou, Constance Owl, Xiaoming Zhai, Ninghao Liu, Claudio Saunt, and Tianming Liu. 2024. Transcending language boundaries: Harnessing llms for low-resource language translation. *Preprint*, arXiv:2411.11295.

Peiqi Sui, Eamon Duede, Sophie Wu, and Richard So. 2024. Confabulation: The surprising value of large language model hallucinations. In *Proceedings of the 62nd Annual Meeting of the Association for Computational Linguistics (Volume 1: Long Papers)*, pages 14274–14284, Bangkok, Thailand. Association for Computational Linguistics.

Khanh-Tung Tran, Barry O'Sullivan, and Hoang Nguyen. 2024a. Irish-based large language model with extreme low-resource settings in machine translation. In *Proceedings of the Seventh Workshop on Technologies for Machine Translation of Low-Resource Languages (LoResMT 2024)*, pages 193–202, Bangkok, Thailand. Association for Computational Linguistics.

Khanh-Tung Tran, Barry O'Sullivan, and Hoang D Nguyen. 2024b. Uccix: Irish-excellence large language model. *arXiv preprint arXiv:2405.13010*.

# A  Appendix A - Irish Verbs

# B  Appendix B - Irish Declensions

| 1st Conjugation | Broad vowels | Slender vowels | 2nd Conjugation | Broad vowels | Slender vowels |
|---|---|---|---|---|---|
| 1st sing | *-aim* | *-im* | 1st sing | *-aím* | *-ím* |
| 2nd sing | *-ann tú* | *-eann tú* | 2nd sing | *-aíonn tú* | *-íonn tú* |
| 3rd sing | *-ann sé/sí* | *-eann sé/sí* | 3rd sing | *-aíonn sé/sí* | *-íonn sé/sí* |
| 1pl | *-aimid* | *-imid* | 1pl | *-aímid* | *-ímid* |
| 2nd pl | *-ann siad* | *-eann sibh* | 2nd pl | *-aíonn sibh* | *-íonn sibh* |
| 3rd pl | *-ann siad* | *-eann siad* | 3rd pl | *-aíonn siad* | *-íonn siad* |
| Aut. | *-tar* | *-tear* | Aut. | *-aítear* | *-ítear* |

Table 16: Suffixes for Conjugation of Irish Verbs in the Present Tense

| Declension | Gender | **Nominative Singular** | **Genitive Singular** |
|---|---|---|---|
| 1st | *M* | *Ends on a broad consonant* | Last consonant is slenderised |
| 2nd | *F (except for im, sliabh)* | *Ends on a consonant either broad or slender* | Ends with '-e' |
| 3rd | *M & F* | *Ends on a consonant either broad or slender* | Ends with '-a' |
| 4th | *M & F* | *Ends with a vowel or '-ín'* | Remains the same as the nominative singular |
| 5th | *F (few M)* | *Ends with '-il', '-in', '-ir' or a vowel* | Ends on a broad consonant |
| **Declension** | Gender | **Nominative Plural** | **Genitive Plural** |
| 1st | *M* | *Same form as the genitive singular* | *Same form as the nominative singular* |
| 2nd | *F (except for im, sliabh)* | *Ends with '-a', e.g. bróga, scornacha* | *Loses the '-a', e.g. bróg, scornach* |
| 3rd | *M & F* | *Ends with '-a', '-acha', '-(a)í', '-(e)anna', '-ta', '-te'* | |
| 4th | *M & F* | *Ends with '-(a)í', '-(e)anna', '-(i)te', '-(i)the', '-nna'* | |
| 5th | *F (few M)* | *Ends with '-(e)acha', '-idí', '-na', '-ne'* | |

Table 17: Verb Declension